\documentclass[conference]{IEEEtran}
\usepackage{mathtools}
\usepackage{graphicx}
\usepackage{caption}
\usepackage{subcaption}
\usepackage{multirow}
\usepackage{float}

\newlength{\qf} 
\graphicspath{{Figures/}{Figures/conf/}}

\begin{document}
\title{Understanding How Image Quality Affects Deep Neural Networks\vspace{-12pt}}
\author{Samuel Dodge and Lina Karam\\
  Arizona State University\\
  sfdodge@asu.edu, karam@asu.edu\vspace{-12pt}}
\maketitle

\begin{abstract}
  Image quality is an important practical challenge that is often overlooked in the design of machine vision systems. Commonly, machine vision systems are trained and tested on high quality image datasets, yet in practical applications the input images can not be assumed to be of high quality. Recently, deep neural networks have obtained state-of-the-art performance on many machine vision tasks. In this paper we provide an evaluation of 4 state-of-the-art deep neural network models for image classification under quality distortions. We consider five types of quality distortions: blur, noise, contrast, JPEG, and JPEG2000 compression. We show that the existing networks are susceptible to these quality distortions, particularly to blur and noise. These results enable future work in developing deep neural networks that are more invariant to quality distortions.
\end{abstract}

\section{Introduction}

Visual quality evaluation has traditionally been focused on the perception of quality from the perspective of human subjects. However, with growing applications of computer vision, it is also important to characterize the effect of image quality on computer vision systems. These two notions of image quality may not be directly comparable as the computer may be fooled by images that are perceived by humans to be identical~\cite{adversarial1}, or in some cases the computer can recognize images that are indistinguishable from noise by a human observer \cite{face-noise}. Thus, it is important to separately consider how image quality can affect computer vision applications. 

In computer vision, recent techniques based on deep neural networks (DNN) have begun to achieve state-of-the-art results in many problem domains \cite{nn-baseline}. Of particular interest to DNN models is image classification performance on large scale datasets with millions of images and thousands of categories. These problem domains were previously thought to be extremely difficult, but DNNs have achieved very impressive results. For example, in the ILSVRC 2010 challenge, the AlexNet DNN\cite{alex} achieved the best result with nearly 9\% better classification accuracy than the second best result based on hand-crafted features.

\begin{figure}[!tb]
  \centering
  \includegraphics[width=0.35\textwidth]{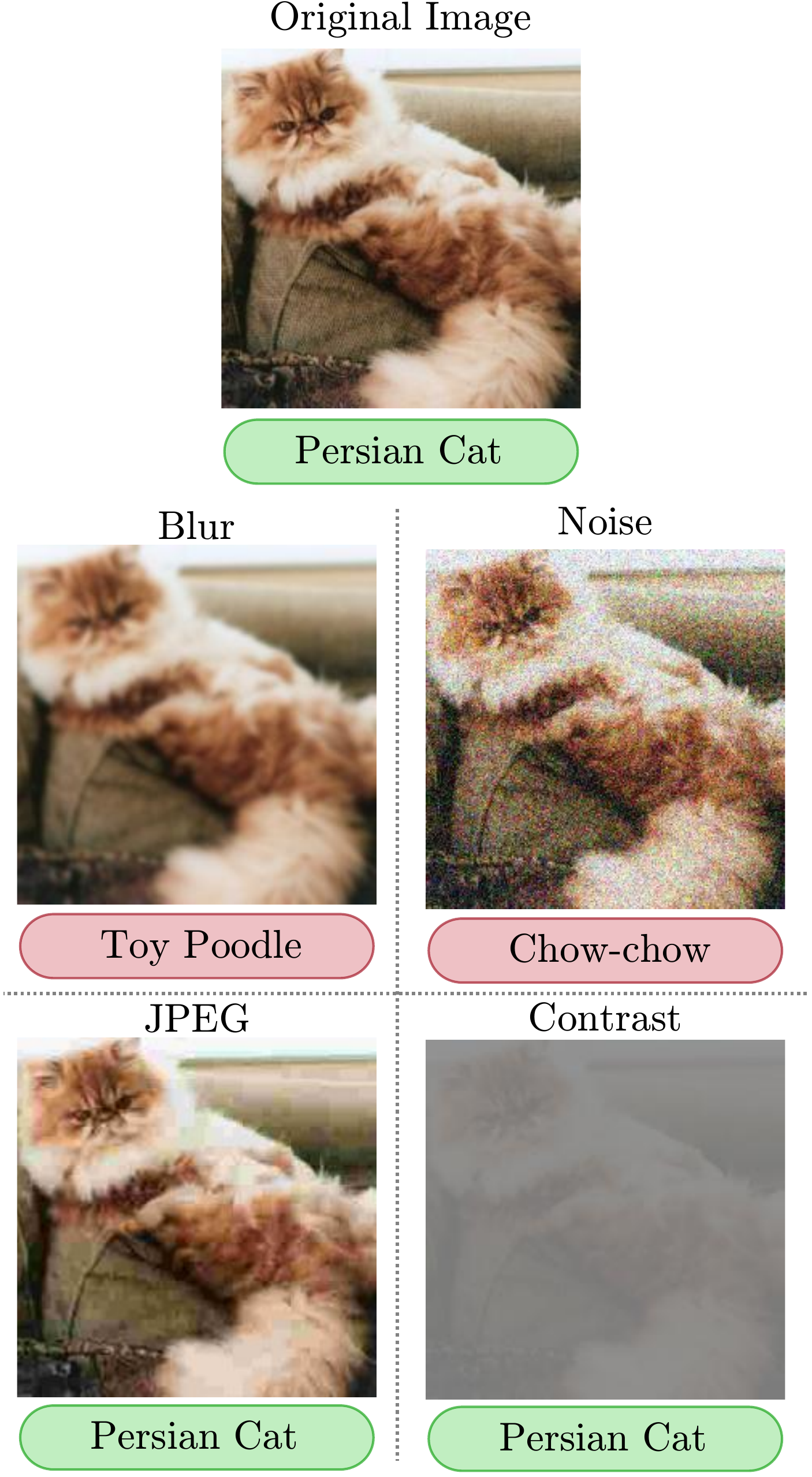}
  \caption{Given the original image, a deep neural network\cite{vgg} correctly identifies the image as ``Persian Cat''. However, when a small amount of Gaussian blur is added ($\sigma = 2$) the network misclassifies the image as ``toy poodle''. Similarly when additive Gaussian noise is added ($\sigma^2 = 50$) the network misclassifies the image as ``chow-chow''. Under JPEG distortion and low contrast the network is able to correctly classify the original image.}
\end{figure}

Despite their impressive performance, deep networks have been shown to be susceptible to adversarial samples \cite{adversarial1}. Adversarial samples are generated by adding worst case noise to the image such that the classification prediction is incorrect with a high confidence. This worst case noise is imperceptible to human observers. The noise is carefully chosen  using either an optimization algorithm \cite{adversarial2} or by exploiting linear properties of the network \cite{adversarial1}. Adversarial samples present an interesting problem, however in practice such carefully chosen noise is unlikely to be encountered. 

It is much more likely that the network will encounter quality distortions stemming from artifacts during image acquisition, transmission, or storage. Even though these sources of noise are not worst case, they can still cause the network to misclassify. For example, Fig. 1 shows the performance of a DNN under several distortions. Under noise and blur the network has difficulty predicting the correct class (``Persian Cat'').

For these types of distortions, we are interested in determining the distortion level at which level the performance begins to decrease. Also, it is interesting to investigate whether the structure of the network significantly affects the ability to be invariant to quality distortions. This will give insight as to what architectures would be useful for building networks that are more invariant to these distortions.

\subsection{Related Works}

For many applications in computer vision it is assumed that the input images are of relatively high quality. However, in certain application domains such as surveillance, image quality is an important consideration. Additionally, with the advent of many computer vision applications on cellular phones, the requirement of high quality images may need to be relaxed.

In surveillance applications, face recognition in low quality images is an important capability. There are many works that attempt to recognize low-resolution faces \cite{face1, face2}. Besides low-resolution, other image quality distortions may affect performance. Karam and Zhu \cite{karam} present a face recognition dataset that considers five different types of quality distortions. They however do not evaluate the performance of any models on this new dataset. Tao et al. \cite{tao} present an approach based on sparse representations that achieves good performance on this dataset.

For hand-written digit recognition, Basu et al. \cite{basu} present the n-MNIST database, which is a modification of the benchmark MNIST dataset. n-MNIST adds Gaussian noise, motion blur, and reduced contrast to the original images. Additionally, the authors in \cite{basu} propose a modification of deep belief networks to achieve greater accuracy on this dataset.  

Ullman et al. \cite{ullman} consider deep neural network performance on low resolution crops of an image. They find minimal recognizable configurations of images (MIRCs) which are the smallest crops for which human observers can still predict the correct class. MIRCs are discovered by repeatedly cropping the input image and asking human observers if they can still recognize the cropped image. The MIRC regions are blurry because in general they represent very small regions. The authors test deep networks on the MIRC regions and show that they cannot match human performance. By contrast, in this paper we consider blurring the entire image rather than selecting a small region of the image, in addition to other types of distortions that occur in practical applications.

In this paper, we present the first large scale evaluation of deep networks on natural images under different types and different levels of image quality distortions. In contrast to \cite{basu,karam}, we use the ILSVRC 2012 dataset (ImageNet) \cite{imagenet} which consists of 1000 object classes. The original images from this database are relatively high quality. We augment this dataset by introducing several distortions and then evaluate the performance of state-of-the-art deep neural networks on these distorted images.

\section{Background}
Here we provide a brief overview of deep neural networks. A more detailed overview can be found in \cite{bengiobook}. Deep neural networks are inspired by biological neural networks. That is, they are a collection of small, simple elements called neurons. In general, a deep network consists of layers of neurons where each neuron computes the following activation function:
\begin{equation}
  f(\mathbf{x}) = \phi (\textbf{w}^T\mathbf{x} + b)
\end{equation}
\vspace{-1pt}\noindent where $\mathbf{x}$ is the input to the neuron, $\mathbf{w}$ is a weight vector, $b$ is a bias term and $\phi$ is a nonlinearity function. Each neuron receives potentially many inputs, and outputs a single number. The nonlinearity is important because it allows layers of neurons to learn non-linear functions. In these layered structures, the output of one layer of units becomes the inputs to the next layer of units. The networks considered in this paper use Rectified Linear Units \cite{relu} as the nonlinearity function.

For image recognition problems, the input to the network is the image itself (usually normalized). However, if a single neuron is to receive inputs from the entire image, the memory and computational requirements quickly become prohibitive. To mitigate this problem weight sharing is used. Rather than each neuron using a separate weight vector $\mathbf{w}$, this vector is shared between neurons. The weight vector connects to nearby neurons from the previous layer within a pre-defined region known as a receptive field. In practice, this process is identical to convolutional filtering with the filter represented by the weights. Layers with convolutional shared weights  are called convolutional layers. Layers without the convolutional shared weights are called fully connected layers.

In addition to convolutional layers, networks often incorporate a max pooling stage. This stage serves two purposes: to improve robustness to noise in filter responses, and to increase the size of the receptive field in the next layer without increasing the size of the filter. This operation considers a window (often 2x2 or 3x3 pixels) and takes the maximum neuron response in each window across the input.

The last stage of the network is typically a softmax layer. The softmax normalizes the responses of the units such that they sum to one. In this way the output layer becomes a probability distribution with each neuron corresponding to the probability (or confidence) of the network for a particular class.

The network parameters ($\mathbf{w}$ and $b$ for each unit) are trained using a large set of input images. First the output of the network is computed for a given set of images. This output is compared to the known class labels and a cost function indicates how closely the predictions match the ground truth. The gradient of this cost function can be computed, and by propagating this gradient backwards through the network, the gradient of each neuron is computed. With the gradient, any number of optimization techniques based on gradient descent can be used to optimize the weights. This general framework is called the backpropagation algorithm \cite{backprop}.

\section{Experimental Setup}

\subsection{Deep Networks}
\begin{table}[!t]
  \centering
 \caption{Summary of DNN models.}
\begin{tabular}{lccc}
  \hline
   Model    & Convolutional & Full & Number \\
   & Layers & Layers & of Parameters \\
  \hline
  Caffe Reference      & 5 & 3  & 61 million     \\
  VGG-CNN-S      & 5 & 3  & ~102 million     \\
  VGG-16      & 13 & 3  & 138 million    \\
  GoogleNet      & 21 (inception layers) & 1  & 7 million     \\
  \hline
  \end{tabular}
\vspace{-12pt}
\end{table}

In this paper we consider four representative neural networks. Table I summarizes the considered models. Although there are many architectures in the literature, the networks we test in this paper represent standard common architectures. These networks have all been trained on the ImageNet dataset \cite{imagenet}. This dataset consists of 1000 classes of images and 1.2 million training images. We use the pre-trained model weights from the Caffe library \cite{caffe}.

The first network we consider is the Caffe Reference Model \cite{caffe}. This is an implementation of the AlexNet network \cite{alex}. The network consists of 5 convolutional layers followed by 3 fully connected layers.

The VGG-CNN-S model \cite{vgg-cnn-s} is similar in structure to the Caffe Reference Model. It also consists of 5 convolutional layers followed by 3 fully connected layers. However this network achieves better performance than the Caffe Reference Model because of slight changes in the layer parameters. For example, the first layer of the Caffe Reference Model uses 48 11 $\times$ 11 filters whereas VGG-CNN-S uses 96 7 $\times$ 7 filters. 

The VGG16 model \cite{vgg} is a much deeper neural network. There are 13 convolutional layers followed by 3 fully connected layers. 

Finally, we consider the GoogleNet model \cite{googlenet}. This model incorporates a type of layer called inception layers. The inception layers process the input with different size filters in parallel and fuse the filter responses together. Because of the inception structure, the network uses far less parameters than the other networks we test. The full structure of the GoogleNet model is more complicated than the previous networks, the details can be found in the original paper\cite{googlenet}.

\begin{figure*}[t]
  \centering
  \includegraphics[width=0.87\textwidth]{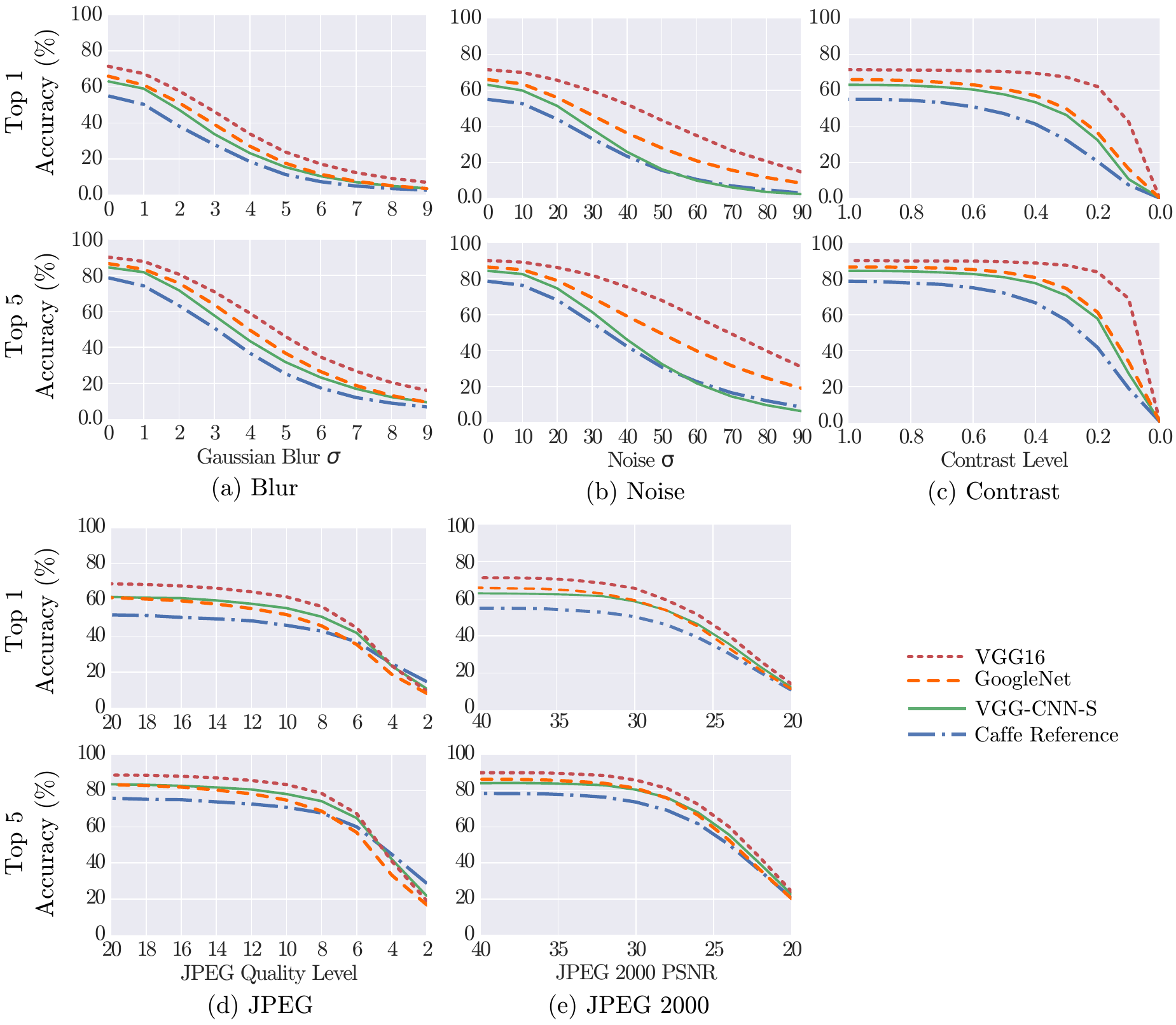}
  \caption{\textbf{Top-1 and Top-5 Accuracy rates under different quality distortions.} The networks are more sensitive to changes in blur and noise compared with compression and contrast.}
\end{figure*}

\subsection{Dataset}
We test on a subset of the validation set of the ImageNet 2012 dataset \cite{imagenet}. To save computation time, we consider 10,000 of the available 50,000 images. We randomly choose 10 images from each of the 1,000 categories. For each image we generate additional images with varying levels of quality distortions.

\subsection{Distortions Types and Levels}

We consider five types of common distortions: JPEG compression, JPEG2000 compression, noise, blur and contrast. We consider each distortion separately. 

Compression is interesting to study because if equivalent performance can be achieved at higher compression ratios, the storage or memory requirement for certain applications can be reduced. For JPEG compression, in our experiments we vary the quality parameter from 2 to 20 in steps of 2. A quality parameter value of 100 represents the original uncompressed image. In initial experiments, we found that the accuracy of the networks does not significantly decrease between quality levels 100 to 20. The LibJPEG library is used to compress the images. For JPEG2000 compression we vary the PSNR from 20 to 40 in steps of 2. The OpenJPEG library is used to compress the images.

Noise may result from using low quality camera sensors. This noise can be modeled as Gaussian noise added to each color component of each pixel separately. We vary the standard deviation of the noise from 10 to 100 in steps of 10.

Blur can result when a camera is not focused properly on the object of interest. Additionally, blur can simulate the network's performance on small or distant objects that will be captured with low resolution. For blur we use a Gaussian kernel and vary the standard deviation of the Gaussian from 1 to 9 in steps of 1. The size of the filter window is set to 4 times the standard deviation.

Finally, we reduce the contrast of the image. Contrast reduction is obtained by blending the input image with a gray image \cite{blend}. The blending factor indicates the level of contrast. We vary the blending factor from 0 to 1 in steps of 0.1.

 \begin{figure*}[!t]
   \centering
   \begin{tabular}{llrrrr}
     \parbox[t]{2mm}{\multirow{5}{*}{\rotatebox[origin=c]{90}{\textbf{Blur}}}}
     & & \includegraphics[width=\qf]{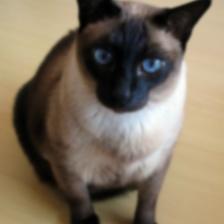}
     & \includegraphics[width=\qf]{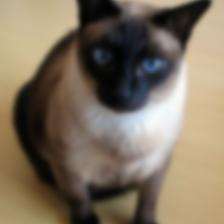}
     & \includegraphics[width=\qf]{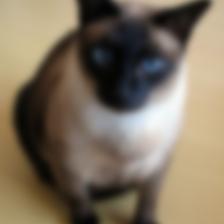}
     & \includegraphics[width=\qf]{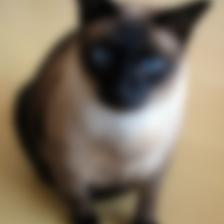} \\
     & Caffe & 0.991843 & 0.991679 & 0.578361  & 0.0305795   \\
     & VGG-CNN-S & 0.998868 & 0.950583 & 0.0315013 & 0.00160614  \\
     & GoogleNet & 0.999726 & 0.993224 & 0.344413  & 0.0950852   \\
     & VGG16 & 0.997842 & 0.985675 & 0.917207  & 0.767373    \\
     \parbox[t]{2mm}{\multirow{5}{*}{\rotatebox[origin=c]{90}{\textbf{Noise}}}}
     & & \includegraphics[width=\qf]{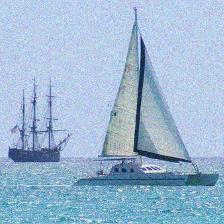}
     & \includegraphics[width=\qf]{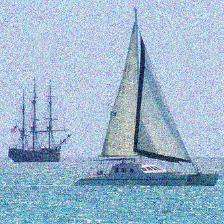}
     & \includegraphics[width=\qf]{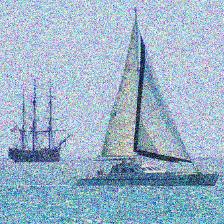}
     & \includegraphics[width=\qf]{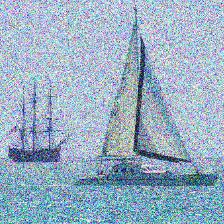} \\
     & Caffe & 0.439129 & 0.496755 & 0.123831  & 0.00186453  \\
     & VGG-CNN-S & 0.354262 & 0.612398 & 0.444991  & 0.0499469   \\
     & GoogleNet & 0.546162 & 0.287545 & 0.130923  & 0.0513721   \\
     & VGG16 & 0.406895 & 0.336332 & 0.48098   & 0.280146    \\
     \parbox[t]{2mm}{\multirow{5}{*}{\rotatebox[origin=c]{90}{\textbf{Contrast}}}}
     & & \includegraphics[width=\qf]{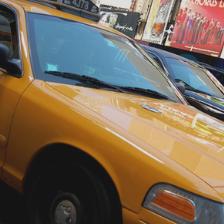}
     & \includegraphics[width=\qf]{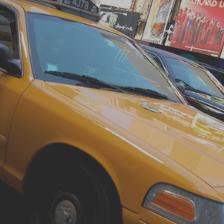}
     & \includegraphics[width=\qf]{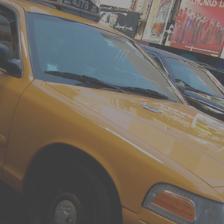}
     & \includegraphics[width=\qf]{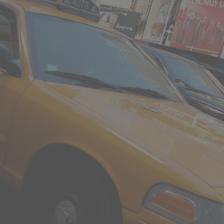} \\
     & Caffe & 0.729299 & 0.635093 & 0.374961  & 0.00598902  \\
     & VGG-CNN-S & 0.921437 & 0.88517  & 0.678366  & 0.0301079   \\
     & GoogleNet & 0.970436 & 0.96616  & 0.871698  & 0.349995    \\
     & VGG16 & 0.834489 & 0.742968 & 0.551311  & 0.32043     \\
     \parbox[t]{2mm}{\multirow{5}{*}{\rotatebox[origin=c]{90}{\textbf{JPEG}}}}
     & & \includegraphics[width=\qf]{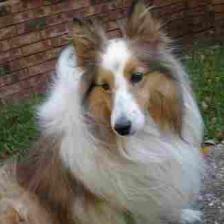}
     & \includegraphics[width=\qf]{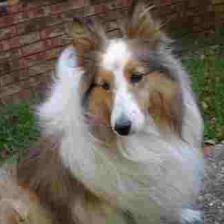}
     & \includegraphics[width=\qf]{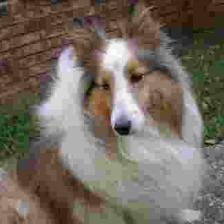}
     & \includegraphics[width=\qf]{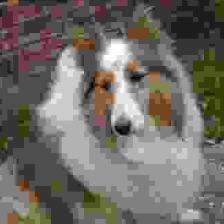} \\
     & Caffe & 0.648897 & 0.628435 & 0.195371  & 0.00520851  \\
     & VGG-CNN-S & 0.661421 & 0.444696 & 0.226847  & 0.000301685 \\
     & GoogleNet & 0.527089 & 0.199268 & 0.11795   & 9.22498e-05 \\
     & VGG16 & 0.817807 & 0.781653 & 0.728399  & 0.000846454 \\
     \parbox[t]{2mm}{\multirow{5}{*}{\rotatebox[origin=c]{90}{\textbf{JPEG 2000}}}}
     & & \includegraphics[width=\qf]{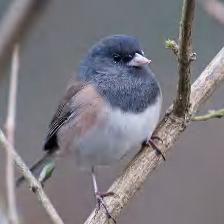}
     & \includegraphics[width=\qf]{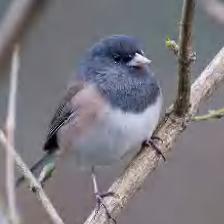}
     & \includegraphics[width=\qf]{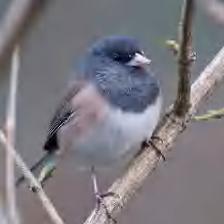}
     & \includegraphics[width=\qf]{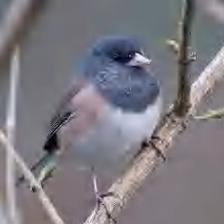} \\
     & Caffe & 0.994832 & 0.961075 & 0.891444  & 0.296613    \\
     & VGG-CNN-S & 0.999976 & 0.999287 & 0.988213  & 0.840988    \\
     & GoogleNet & 0.999978 & 0.997871 & 0.995659  & 0.961735    \\
     & VGG16 & 0.999402 & 0.972143 & 0.862521  & 0.357437    \\
     \end{tabular}
   \caption{\textbf{Example distorted images.} For each image we also show the output of the soft-max unit for the correct class. This output corresponds to the confidence the network has of the considered class. For all networks and for all distortions this confidence decreases as the image quality decreases.}
\end{figure*}

\section{Results}

\begin{figure*}[ht]
  \centering
  \includegraphics[width=0.88\textwidth]{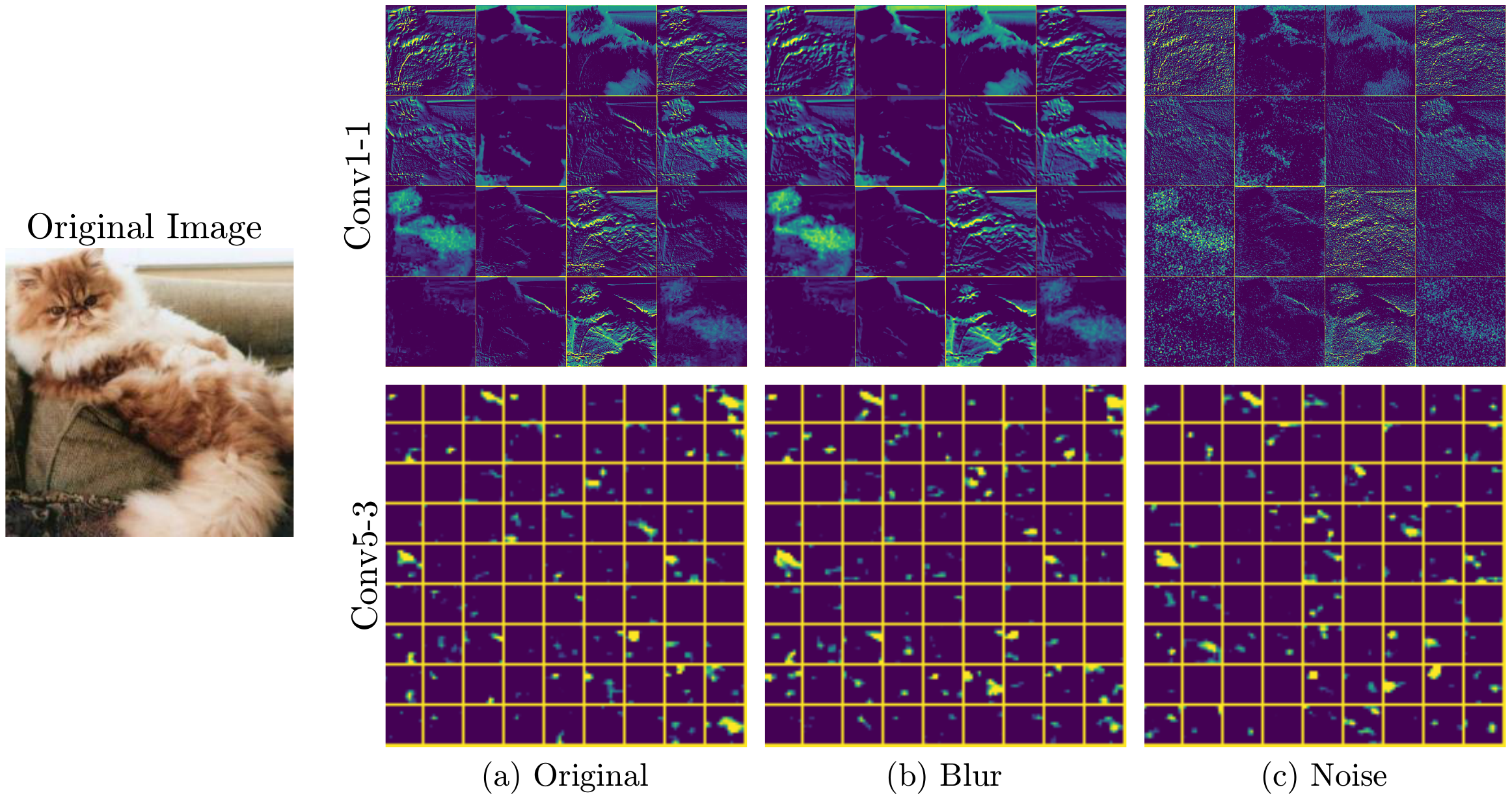}
  \caption{Filter outputs from the VGG16 network on the first (conv1-1) and last (conv5-3) convolutional layers. Blur does not significantly affect the early filter responses, yet the filter responses at the last layer can be quite different as compared to the last layer of the original undistorted image. Noise gives rise to many activations in the first layer, which also propagate to the responses in the last layer.}
\end{figure*}

We consider two measures of accuracy: top-1 classification accuracy and top-5 classification accuracy. The output of the network is a probability for each class. These probabilities can be sorted to give a list of predicted classes with decreasing confidence. The top-1 accuracy measures the accuracy by comparing the top prediction with the correct class. The top-5 accuracy labels a prediction as correct if the correct class is in the top 5 predicted classes. The reason that top-5 accuracy is often reported in the literature is that for some images in the dataset there are multiple objects in the image. The ground truth label for the image is typically the dominant object in the image. The top-5 accuracy allows the network to predict one of the less dominant objects in the image.

Fig. 2 shows the results of our experiments. All of the networks are very sensitive to blur. Even for moderate blur levels, the accuracy of the networks decreases significantly. This reduction in performance can possibly be explained because blur removes textures in these images. The network may be looking for specific textures to classify an image.

Images under noise exhibit a similar decrease in performance. However, compared with the Caffe network and the VGG-CNN-S network, the performance of the VGG-16 and GoogleNet falls off  slower. This could be because of the deeper structure of the VGG-16 and GoogleNet networks, which allows the networks more room to learn features that are not affected by noise. At a noise standard deviation of 90 the networks performance becomes less than 20\% on average; however at this level of noise, the images are still easily recognizable by human observers.

The networks are surprisingly resilient to JPEG and JPEG2000 compression distortions. It is only at very low quality levels (quality parameter less than 10 for JPEG and PSNR less than 30 for JPEG2000) that the performance begins to decrease. This means that we can be reasonably confident that deep networks will perform well on compressed data, given that the compression level is sufficient.


The networks also show resiliency with respect to contrast changes. Again, the VGG-16 network appears more robust than the other networks. 

In Fig. 3 we show a closer look at several individual test images. For each image we show the soft-max unit output for the correct class. Numbers closer to 1 mean that the network is very confident of the prediction, and numbers close to 0 imply little confidence. We see that the networks become less confident as the quality level decreases.

In order to understand why the performance of the networks is sensitive to blur and noise, we examine the filter outputs from the network under blur and noise separately. In Fig. 4 we examine the response for the Persian Cat image from Fig.~1.

The blur causes the filter responses in the first convolutional layer to change, but only slightly. However, at the last convolutional layer, the filter responses exhibit significant changes when blur is present as compared to the responses generated using the original undistorted images. This tells us that small changes in the first layer response are propagating to create larger changes at the higher layer. 

The effect of noise is more evident in the first convolutional layer. The high frequency nature of the noise is picked up by many of the early layer filters. The responses of the last convolutional layer are not so noisy, but are much different than those from the original image.

\section{Discussion and Conclusion}
Our results show that of the neural networks tested, all are susceptible to blur and noise distortions, while being resilient to compression artifacts and contrast. This is an interesting result because it shows that the reduced performance under blur and noise is not limited to a particular model, but is common to the considered DNN architectures. To create networks that are invariant to these properties, new designs may need to be introduced.

One obvious solution to this problem is to train the networks on low quality images. In fact, the Caffe and GoogleNet are already trained with slight noise (in the form of color perturbations) added to the image to add regularization. Despite this, the models still perform poorly. Training with low quality images should improve testing results on low quality images, but perhaps this may also result in decreased performance on high quality images. Furthermore, training with increased number of samples leads to longer training times. An investigation of the benefits of training with low quality samples is left for future work.

Our results also show that the VGG-16 network\cite{vgg} exhibits the best performance in terms of classification accuracy and resilience for all types and levels of distortions as compared to the other networks.

\vspace{-8pt}
\section*{Acknowledgment}
The authors would like to thank NVIDIA Corporation for the donation of a TITAN X GPU.

\bibliographystyle{IEEEtran}
\bibliography{refs}
\end{document}